\pdfoutput=1

\documentclass[11pt]{article}

\usepackage[final]{acl}

\usepackage{times}
\usepackage{latexsym}
\usepackage{booktabs}
\usepackage{multicol}
\usepackage{multirow}
\usepackage{subcaption} 
\usepackage{color, colortbl}

\definecolor{lightgray}{gray}{0.9}
\definecolor{lightblue}{rgb}{0.93,0.95,1.0}
\definecolor{darkgreen}{rgb}{0.0,0.6,0.0}
\definecolor{darkblue}{rgb}{0.0,0.0,0.5}
\definecolor{pinegreen}{rgb}{0.0, 0.47, 0.44}
\definecolor{deepmagenta}{rgb}{0.8, 0.0, 0.8}
\definecolor{amber}{rgb}{1.0, 0.49, 0.0}
\definecolor{bananayellow}{rgb}{1.0, 0.88, 0.21}
\definecolor{citrine}{rgb}{0.89, 0.82, 0.04}
\newcommand\orange[1]{\textcolor{amber}{\textbf{#1}}}

\newcommand\yellow[1]{\textcolor{citrine}{\textbf{#1}}}
\newcommand\green[1]{\textcolor{forestgreen}{\textbf{#1}}}
\newcommand\blue[1]{\textcolor{blue}{\textbf{#1}}}

\newcommand{\smodel}{TraveLER}
\usepackage{amsmath}
\usepackage{mdframed}
\usepackage{soul}

\def\Secref#1{Section~\ref{#1}}

\newcommand{\minisection}[1]
{\noindent{\textbf{#1}.}}
\newcommand{\tabref}[1]{Table~\ref{#1}}

\newcommand{\figgref}[1]{Figure~\ref{#1}}

\newcommand{\tablestyle}[2]{\setlength{\tabcolsep}{#1}\renewcommand{\arraystretch}{#2}\centering\footnotesize}
\newlength\savewidth

\newcommand{\gcol}[1]{{\bf \fontsize{6.5}{42}\selectfont \color{citecolor!80}~(#1)}}

\definecolor{citecolor}{RGB}{34,139,34}
\definecolor{lightred}{RGB}{241,140,142}
\definecolor{amber(sae/ece)}{rgb}{1.0, 0.49, 0.0}
\definecolor{battleshipgrey}{rgb}{0.52, 0.52, 0.51}
\definecolor{cadmiumorange}{rgb}{0.93, 0.53, 0.18}
\definecolor{applegreen}{rgb}{0.55, 0.71, 0.0}
\definecolor{cadmiumgreen}{rgb}{0.0, 0.42, 0.24}
\definecolor{forestgreen}{rgb}{0.13, 0.55, 0.13}
\definecolor{red}{rgb}{0.89, 0.0, 0.13}
\definecolor{lavendermist}{rgb}{0.9, 0.9, 0.98}

\usepackage{adjustbox}
\usepackage{multirow}
\usepackage{tablefootnote}
\usepackage{color, colortbl}
\usepackage{tcolorbox}
\usepackage{svg}
\usepackage{subcaption}

\definecolor{lavendermist}{rgb}{0.9, 0.9, 0.98}
\definecolor{lightgray}{gray}{0.9}

\usepackage[T1]{fontenc}

\usepackage[utf8]{inputenc}

\usepackage{microtype}

\usepackage{inconsolata}

\usepackage{graphicx}

\title{TraveLER: A Modular Multi-LMM Agent Framework \\for Video Question-Answering}

\author{
\textbf{Chuyi Shang\textsuperscript{$\star$}~~~}
\textbf{Amos You\textsuperscript{$\star$}~~~}
\\
\textbf{Sanjay Subramanian}~~~
\textbf{Trevor Darrell}~~~
\textbf{Roei Herzig}
\\
\\
University of California, Berkeley
}

\begin{document}
\maketitle
\begin{abstract}
Recently, image-based Large Multimodal Models (LMMs) have made significant progress in video question-answering (VideoQA) using a frame-wise approach by leveraging large-scale pretraining in a zero-shot manner. Nevertheless, these models need to be capable of finding relevant information, extracting it, and answering the question simultaneously. Currently, existing methods perform all of these steps in a single pass without being able to adapt if insufficient or incorrect information is collected. 
To overcome this, we introduce a modular multi-LMM agent framework based on several agents with different roles, instructed by a Planner agent that updates its instructions using shared feedback from the other agents.
Specifically, we propose \textit{TraveLER}, a method that can create a plan to ``\textbf{Trave}rse'' through the video, ask questions about individual frames to ``\textbf{L}ocate'' and store key information, and then ``\textbf{E}valuate'' if there is enough information to answer the question. Finally, if there is not enough information, our method is able to ``\textbf{R}eplan'' based on its collected knowledge. Through extensive experiments, we find that the proposed \textit{TraveLER} approach improves performance on several VideoQA benchmarks without the need to fine-tune on specific datasets. Our code is available at \href{https://github.com/traveler-framework/TraveLER}{https://github.com/traveler-framework/TraveLER}.
\end{abstract}

\renewcommand*{\thefootnote}{*}
\footnotetext{Equal contribution.}
\renewcommand*{\thefootnote}{}

\vspace{-5pt}
\begin{figure*}[t!]
\href{https://youtu.be/n_KGHYOE77w}{\includegraphics[width=0.95\linewidth]{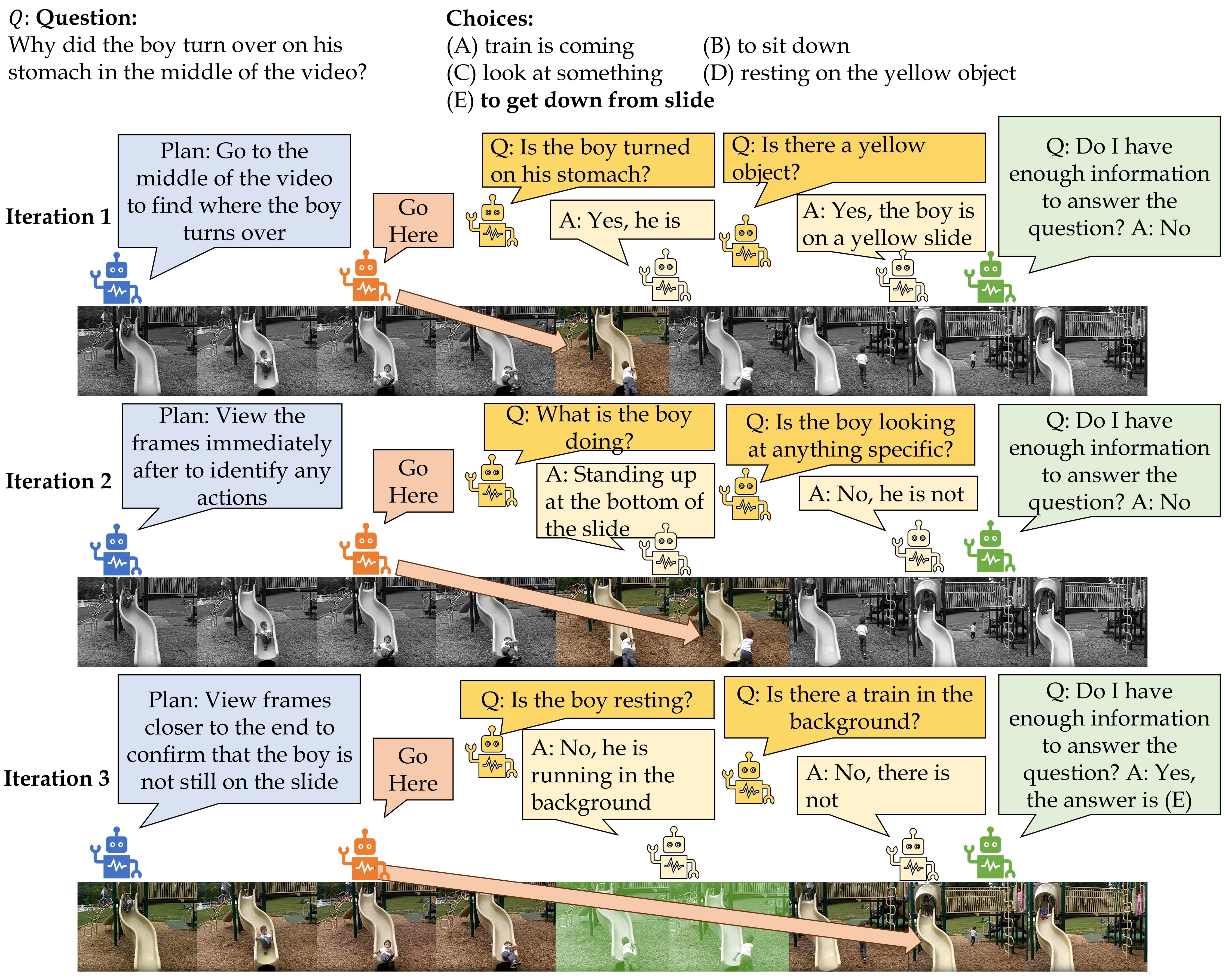}}
  \caption{\textbf{A simplified overview of our {\smodel} framework.} Our proposed framework aims to answer the question by collecting relevant information from keyframes through interactive question-asking. To accomplish this, several agents (in colored boxes) with different roles interact (left-to-right in each row) over several iterations. {\smodel} creates a plan (in \blue{blue}) to ``traverse'' (in \orange{orange}) through the video, asks questions regarding individual frames (in \yellow{yellow}) to ``locate'' and store key information and, ``evaluates'' whether there is sufficient information to answer the question (in \green{green}), and ``replans'' using past collected knowledge if there is not enough information. \textbf{Click on the image to see the video}.}
    \label{fig:teaser}
\end{figure*}

\section{Introduction}
\label{sec:intro}

Over the last few years, Large Multimodal Models (LMMs) have demonstrated tremendous progress in the area of video understanding, particularly for the video question-answering (VideoQA) domain~\citep{fu2021violet,wang2022internvideo}. More recently, LMMs have been able to achieve impressive results through video-based models~\citep{lin2023videollava,NEURIPS2022_f8290cc, ye2023hitea, li2023lavender}. However, video models require a high level of computational complexity to fine-tune, and annotations are difficult and expensive to collect. As a result, many recent approaches~\citep{xue2023clipvip, yu2023self, zhang2023simple} operate on the frame level, leveraging large-scale image-based pretrained models in a zero-shot manner. 

Moreover, these models may need to do several tasks simultaneously in a single step. In particular, they should identify the correct events in videos by understanding what information is relevant and ignoring irrelevant information. Next, they would need to extract specific and question-relevant visual details and use them to answer the question. As such, one iteration might not be enough to collect all the necessary information. For example, many current approaches use simple captioning, which is often too general to extract specific details, or they might miss important events. In these cases, current approaches cannot revisit the video to find additional information. Decomposing this process into different components allows each component to adapt to newly collected information.

To address this, we introduce a modular multi-LMM agent framework for VideoQA. A Planner agent instructs agents in charge of different tasks, such as navigating through the video, extracting visual information through a question-answering process, and reviewing this information to select an answer. Feedback from the agents is then passed on to each other and back to the Planner, who uses the feedback to update its instructions. In this way, we can revisit the video to narrow our focus or expand our search to more relevant information and extract specific details to answer the question.


Consider the example in~\figgref{fig:teaser}. Suppose we are asked why the boy turned over in the middle of the video. In the first iteration, our method uses temporal cues from the question to skip to the middle of the video and asks questions to find the relevant frames. In the next iteration, we gather more information. Asking about what the boy is doing, we learn that he is ``standing up at the bottom of the slide" and is not looking at anything specific, which informs us that the boy is no longer ``sitting down" (choice B) or ``resting on the yellow object" (choice D). To eliminate these choices, we must confirm that the boy does not sit back down again by traveling to a timestamp near the end of the video. Finally, since we have collected enough information and followed the plans, we can select the right choice that the boy turns over to be on his stomach ``to get down from slide'' (Choice E).

Our proposed approach -- \textbf{Trave}rse, \textbf{L}ocate, \textbf{E}valuate, and \textbf{R}eplan (\smodel), has four main stages. First, in the Traversal stage, an agent creates a plan to answer the question. In the Location stage, an agent uses the plan to decide which timestamp of the video to select. The corresponding frames are then sent to another agent, which asks questions and stores the answers in a memory bank for future iterations. Finally, in the Evaluation stage, an agent reviews all collected information and decides whether to answer or create a modified plan (Replan) to start the next iteration if necessary.

To summarize, our main contributions are:
(i) We introduce {\smodel}, a modular multi-LMM agent framework for video question-answering. (ii) Our method shows improved performance on multiple difficult video question-answering benchmarks, such as NExT-QA, EgoSchema, Perception Test, and STAR. (iii) Our method is easy to employ with different LLMs and LMMs, highlighting the effectiveness of our modular approach.
\section{Related Work}

\minisection{Video Question-Answering} VideoQA involves answering free-form or multiple-choice questions given an input video. Compared to image question answering, VideoQA poses unique challenges because it often requires strong temporal understanding and the ability to deal with long input sequences. Many recent works have focused on training end-to-end video-language models \citep{fu2021violet, NEURIPS2022_f8290cc, wang2022internvideo, ye2023hitea, yu2022coca, li2023lavender}, but doing so remains challenging due to computational constraints and difficulties in architecture scaling. As a result, many approaches adopt pretrained image models to the video domain by extracting information independently from each frame \citep{xue2023clipvip, yu2023self, zhang2023simple}. Here, we design a framework that builds an adaptive plan to find and extract relevant information using a question-answering approach.

\minisection{LMMs for Video Understanding}
LMMs have been shown to be extremely useful for VideoQA. Some methods use supervised or contrastive training to perform video-LMM pretraining \citep{zhao2023lavila, yang2023vid2seq, chen2024vast}, while others adapt existing LMMs and use instruction tuning to adapt them to the video domain~\citep{zhang2023videollama, maaz2023videochatgpt, lin2023videollava}. However, recent improvements in LMM capabilities have allowed for many strong approaches for few-shot~\citep{alayrac2022flamingo, NEURIPS2022_381ceeae} and zero-shot VideoQA \citep{yang2022frozenbilm}. In particular, zero-shot methods, such as LLoVi~\citep{zhang2023simple}, use pre-trained LMMs to generate captions for each frame in the video. Nevertheless, uniformly sampling frames at random may miss important visual information and focus on unimportant frames ~\citep{Wu_2019_ICCV, lei2021less}. Recent works like SeViLA~\citep{yu2023self} addressed this problem by performing parameter-efficient finetuning using captions to identify keyframes~\citep{lu2022lgdn, buch2022revisiting, qian2023locate}, but this requires fine-tuning on specific datasets. Unlike these works, which select all keyframes in a single pass, we introduce a novel iterative and modular approach instructed by a planner.


\minisection{LMM-based Agents for Videos}
The strong reasoning abilities of LLMs ~\citep{brown2020language, chung2022scaling} have made them effective in LLM-based agent approaches for videos, where an LLM performs much of the reasoning after collecting information from different modules~\citep{chen2023video, lin2023mmvid, zhang2023mmnarrator, zeng2022socraticmodels}. For example, Socratic Models~\citep{zeng2022socraticmodels} proposes a method to reason about videos based on generated audio transcriptions and CLIP frame similarity scores, while other works like VideoChatCaptioner~\citep{chen2023video} proposes a way to caption videos through chat dialogues between an LLM and a LMM. Recently, there have also been works that use program generation using an LLM to answer questions ~\citep{choudhury2023zeroshot, min2024morevqa}. However, these works still use a single-pass approach and provide very general captions. In contrast, our work uses an iterative question-answering process to extract specific, relevant information in the frame.


\begin{figure*}[tb!]
  \centering
  \includegraphics[width=0.95\linewidth]{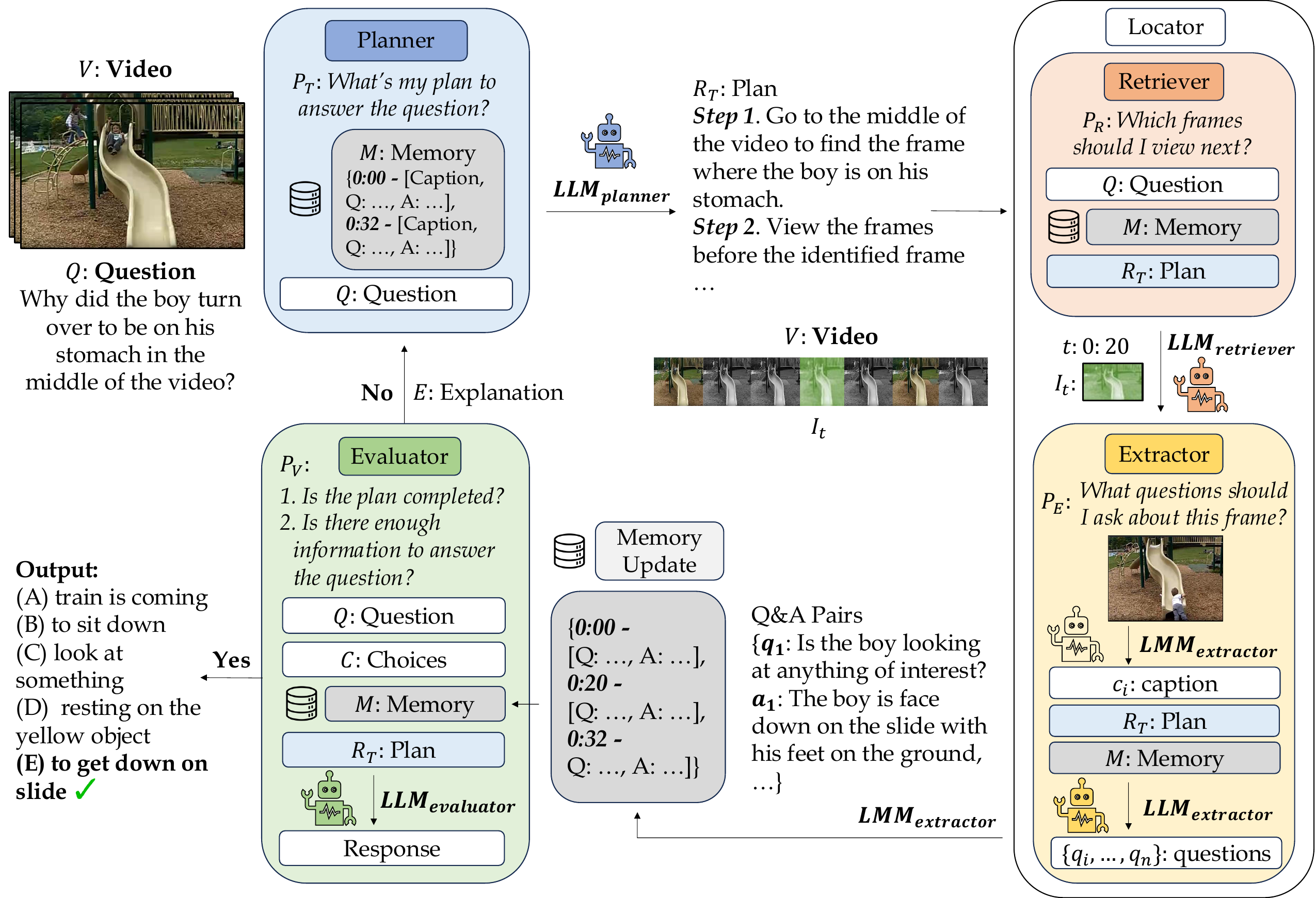}
  \caption{\textbf{TraveLER framework.} Our framework consists of four different modules, the Planner, Retriever, Extractor, and Evaluator. The Planner creates a plan and sends it to the Retriever. The Retriever uses the plan to select the next timestamp and sends this to the Extractor. The Extractor captions and generates questions about the timestamp, answers the questions, and saves the output in the memory bank. Finally, the Evaluator determines if there is enough information and if the plan has been followed. If yes, the Evaluator returns the answer, else the existing information is sent back to the Planner to begin a new iteration.}
    \label{fig:arch}
\end{figure*}

\section{TraveLER Framework}
\label{sec:method}


We begin by describing some background on the LLM and LMM architectures (\Secref{sec:method:preliminaries}), then introduce each component of our framework (\Secref{sec:method:TraveLER}) and implementation details (\Secref{sec:method:impl}). Our method is shown in~\figgref{fig:arch}.

\subsection{Preliminaries}
\label{sec:method:preliminaries}
\minisection{Large language and multimodal models} LLMs are text-conditioned generative models. Given a prompt $P$, they encode it into a fixed language embedding $l$ in an embedding space $f(\cdot)$ and use this to produce text response $R$: $R = f(l(P))$. Similarly, Large Multimodal Models (LMMs) are adapted to jointly reason over vision and language. To map different modalities into the shared embedding space $f(\cdot)$, an image $I$ is encoded using an encoder $v$, and the prompt $P$ is encoded using a fixed language embedding $l$. The LMM outputs a textual response $R$:  $R = f(v(I), l(P))$.

\minisection{Video question-answering}
VideoQA involves viewing a video and answering questions. The model is usually evaluated through \textit{top-1} accuracy, which chooses the best answer out of a set of possible choices. Specifically, given a question $Q$, video input $V$ consisting of a set of frames $\{I_1, \cdots, I_n\}$, and a set of choices $C = \{c_1, \cdots, c_n\}$, the model is asked to choose the best $c_i$ to answer $Q$. Next, we introduce each component of our method.

\subsection{TraveLER Components}
\label{sec:method:TraveLER}

\minisection{Traversal} In the Traversal stage, we create a plan for how to traverse through the video, which is a list of textual instructions that guide our approach to answering the question. To achieve this, we use the task prompt $P_T$, which is an instruction to create a plan for answering the question. We combine $P_T$ with the question $Q$, and memory bank $M$, which is a dictionary of collected information keyed by timestamps and containing information from the corresponding frame, to receive the final prompt $P_\mathrm{T}^{(1)}$: $P_\mathrm{T}^{(1)} = ``[Q][M][P_T]"$. 

Our method uses a memory bank $M$ to store collected information, which allows information to persist and to be updated as we proceed through different iterations. We initialize $M$ with captions of 5 evenly sampled frames throughout the video. We find that this memory initialization gives the model good context for the general idea of the video, and performs better than starting with an empty memory $M$. After the first iteration, we add information iteratively using the Extractor module, which we discuss later in this section.

Next, we input the prompt $P_\mathrm{T}^{(1)}$ into $\text{LLM}_{\text{planner}}$, which returns response $R_T$, a step-by-step plan on how to traverse through the video and what information should be collected. This plan is revised in future iterations using collected information.
\nolinebreak
$$R_T = \text{LLM}_{\text{planner}}(l(P_\mathrm{T}^{(1)}))$$
\nolinebreak
Our next step is to use the plan $R_T$ in the Locator stage to locate relevant events and extract the information that we will use to answer the question.

\minisection{Locator} The Locator is a component that consists of two submodules, the \textit{Retriever} and the \textit{Extractor}. The Retriever selects the timestamps of the next frames to view, while the Extractor extracts relevant information from these frames using a question-answering process. Next, we discuss each component in more detail.

\textit{(i) Retriever:} The Retriever carries out the given plan $R_T$ by selecting which frames to view next. The Retriever is an LLM-based submodule that finds the next timestamp $t$ to view, given the plan $R_T$, collected information $M$, and video metadata (frame rate, length). The task prompt $P_R$ is an instruction that contains information about the video length and asks which timestamp to view next. 
Thus, we insert the question $Q$, the plan $R_T$, and the collected information $M$ into the task prompt $P_R$ to create the new prompt $P_\mathrm{R}^{(1)}$: $P_\mathrm{R}^{(1)} = ``[P_R][Q][R_T][M]"$.

Given prompt $P_\mathrm{R}^{(1)}$, the LLM in the Retriever, $\text{LLM}_{\text{retriever}}$, returns $t$, the next set of timestamps. Then, it retrieves frames $I_t$ at timestamp $t$.

\textit{(ii) Extractor:}
The Extractor is important because it allows us to capture more relevant and question-specific details from the visual input, unlike using only captions.
We pass the frames selected by the retriever $I_t$ into the Extractor submodule, which consists of two large models: $\text{LLM}_{\text{extractor}}$, to generate context-dependent questions about the frames $I_t$, and a different vision-language model $\text{LMM}_{\text{extractor}}$, which extracts the desired information from the same frames. We note that we use both an LLM and LMM since the LLM is better at reasoning about the plan and the collected information, while the LMM is able to collect visual information requested by the LLM.

We first generate a general caption $c_t$ for frame $I_t$ using the $\text{LMM}_{\text{extractor}}$. Then, we concatenate the caption $c_t$, plan $R_T$, and memory $M$, and the Extractor task prompt $P_E$, which is an instruction that asks to use available information to create 3 questions to ask about the current frame. This results in the new prompt $P_\mathrm{E}^{(1)}$: $P_\mathrm{E}^{(1)} = ``[P_E][c_t][R_T][M]"$.

Next, we input this new prompt $P_\mathrm{E}^{(1)}$ into the LLM to get a set of questions $\{q_1, q_2, \cdots, q_n\}$ about each frame, where $n$ is a parameter for how many questions to ask about each frame. 
\nolinebreak
$$\{q_1, q_2, \cdots, q_n\} = \text{LLM}_{\text{extractor}}(l(P_\mathrm{E}^{(1)}))$$
\nolinebreak
\noindent where $l$ is the fixed language embedding.

In this way, the generated questions take into account both the plan $R_T$ and information from past and future frames of the video $M$. We then use the frame $I_t$, and the corresponding questions $\{q_1, q_2, \cdots, q_n\}$ as input into $\text{LMM}_{\text{extractor}}$. The $\text{LMM}_{\text{extractor}}$ then outputs a set of answers $\{a_1, a_2, \cdots, a_3\}$, where each answer $a_i$ corresponds to the question $q_i$.
\nolinebreak
$$\{a_1, \cdots, a_n\} = \text{LMM}_{\text{extractor}}(v(I_t), l(\{q_1, \cdots, q_n\}))$$

\noindent where $v$ is the visual encoder.

Finally, to use this collected information in future iterations, we update our memory bank $M$. To do this, we use the timestamp $t$ of $I_t$ as our key and the question-answer pair list as the value, and append this to our memory $M$. If the memory bank dictionary $M$ is too long, we summarize it by using the memory bank as an input to another LLM and instruct it to make the memory bank entries more concise, while retaining the same keys and format. This output becomes our new memory bank.

\minisection{Evaluator}
The Evaluator decides if there is enough information and determines if the plan has been followed. We concatenate the memory information $M$, the plan $R_T$, the question $Q$, and the choices $C$ with the task prompt $P_V$, The task prompt $P_V$ is an instruction to evaluate if there is enough information to answer the question and if the given plan $R_T$ has been fulfilled. Thus, we get the new prompt $P_\mathrm{V}^{(1)}$: $P_\mathrm{V}^{(1)} =``[P_\mathrm{V}][Q][C][R_T][M]"$.

We use this prompt $P_\mathrm{V}^{(1)}$ as input into the LLM in the Evaluator, $\text{LLM}_{\text{evaluator}}$, which evaluates if there is enough information to answer the question and if the plan has been completely followed. If both are true, $\text{LLM}_{\text{evaluator}}$ outputs the best choice $c^*$ to answer the question $Q$. Otherwise, it provides an explanation $E$ on why there is insufficient information and gives this explanation to the Planner to start a new iteration of the process.

\minisection{Re-planning}
After each iteration, if the evaluator decides that there is not enough information to answer the question $Q$ or if the plan $P$ has not been completed, the existing memory $M$ will be provided to the Planner in the next iteration, in addition to an explanation $E$ for why an answer was not chosen. The Planner then outputs a new plan, restarting the process. We also set a limit on the number of iterations a question can take to prevent infinite loops. After reaching this limit, we force the Evaluator to choose the best choice.

\minisection{Summarizer}
The Summarizer is an optional module used for some datasets to summarize the information. It is given the question $Q$, choices $C$, memory bank $M$, and task prompt $P_S$, which is an instruction to summarize the information in each timestamp of the memory bank $M$. The outputted summary is then used to replace the original memory bank $M$. This process helps reduce the amount of information passed to the LLM in future steps, which may struggle with very long inputs.

\renewcommand{\thefootnote}{\fnsymbol{footnote}}
\begin{table*}[tb!]
    \label{tab:experiments}
    \caption{\textbf{Results on Datasets.} We show zero-shot results on different datasets.  For fair comparisons, we \textcolor{gray}{gray out} methods with fine-tuned components in their model.} 
    \centering
    \begin{subtable}[t]{.33\linewidth}
	\tablestyle{3pt}{1.25}
        \centering
        \caption{{NExT-QA}}
        \label{tab:nextqa}
        \small
\begin{tabular}{lccccc}
    \toprule
    Model & Cau. & Tem. & Des. & Avg.\\
    \midrule
    \rowcolor{gray!20} SeViLA & 61.3 & 61.5 & 75.6 & {63.6} \\
    \rowcolor{gray!20} MC-ViT-L & - & - & - & 65.0  \\
    \midrule
    InternVideo & 43.4 & 48.0 & 65.1 & 49.1 \\
    ViperGPT & - & - & - & 60.0 \\
    ProViQ & - & - & - & 63.8 \\
    LLoVi & 69.5 & 61.0 & 75.6 & 67.7 \\
    \rowcolor{lavendermist}
    TraveLER& \textbf{70.0} & 60.5 & \textbf{78.2} & \textbf{68.2}\gcol{+0.5} \\
    \bottomrule
\end{tabular}
    \end{subtable}
    \hfill
    \begin{subtable}[t]{.2\linewidth}
	\tablestyle{3pt}{1.25}
        \centering
        \caption{{EgoSchema (Full)}}
        \label{tab:egoschema}
        \small
\begin{tabular}{lc}
    \toprule
    Model & Acc.\\
    \midrule
    mPLUG-Owl & 31.1 \\
    InternVideo & 32.1 \\
    LongViViT & 33.3 \\
    Vamos & 48.3 \\
    LLoVi & 50.3 \\
    \rowcolor{lavendermist}
    TraveLER & \textbf{53.3}\gcol{+3.0} \\
    \bottomrule
\end{tabular}
    \end{subtable}
    \hfill
    \begin{subtable}[t]{0.20\linewidth}
        \tablestyle{3pt}{1.25}
        \centering
	    \caption{{STAR}}
        \label{tab:star}
	\small
\begin{tabular}{cccc}
    \toprule
    Model & Avg. \\
    \midrule
    \rowcolor{gray!20} SeViLA & 44.6 \\
    \midrule
    Flamingo-9B & 41.8 \\
    InternVideo & 41.6\\    
    BLIP-2\textsuperscript{voting} & 40.3\\
    BLIP-2\textsuperscript{concat} & 42.2\\
    \rowcolor{lavendermist} TraveLER & \textbf{44.9} \gcol{+2.7} \\
    \bottomrule
\end{tabular}
    \end{subtable}
    \hfill
    \begin{subtable}[t]{0.20\linewidth}
	\tablestyle{3pt}{1.25}
        \centering
        \caption{{Perception Test}}
        \label{tab:perceptiontest}
        \small
\begin{tabular}{lc}
    \toprule
    Model & Acc. \\
    \midrule
    \rowcolor{gray!20} SeViLA & 46.2 \\
    \rowcolor{gray!20} MC-ViT-B & 47.0  \\
    \rowcolor{gray!20} MC-ViT-L & 48.1 \\
    \midrule
    Flamingo-3B & 43.6 \\
    LongViViT & 45.7\\
    \rowcolor{lavendermist} TraveLER & \textbf{50.2} \gcol{+4.5} \\
    \bottomrule
\end{tabular}
    \end{subtable}
    \label{table:results}
\end{table*}

\renewcommand{\thefootnote}{\arabic{footnote}}

\subsection{Implementation Details}
\label{sec:method:impl}

Here, we discuss how we implement various components of our framework. More implementation details, such as prompts and dataset-specific details are in the Supplementary in~\Secref{supp:impl}.

\minisection{Memory bank}
We represent past collected information as a Python dictionary, with the timestamp of different frames as keys and a list of extracted information from the frame as the values. This extracted information consists of a brief caption of the frame and a list of question-answer pairs. To prevent the memory bank from becoming too large, we also implement a summarizer module that instructs an LLM to summarize the memory bank and return a more concise version in the same dictionary format as before.

\minisection{Agent model selection} Our modular approach has the benefit of allowing us to easily swap in different LLMs and LMMs (see \Secref{sec:eval:ablations}). For our main experiments, we use LLaVA-1.6 \citep{liu2023llava} for $\text{LMM}_{\text{extractor}}$ and GPT-3.5/GPT-4 \citep{openai2023gpt4} for $\text{LLM}_{\text{planner}}$, $\text{LLM}_{\text{retriever}}$, $\text{LLM}_{\text{extractor}}$, and $\text{LLM}_{\text{evaluator}}$.

\minisection{Multi-frame selection}
We also allow the Retriever to select multiple frames instead of one. This helps to capture better events that happen quickly or require more context to recognize. For example, if we want to find the action of "a woman clapping her hands", a single frame selection may cause us to incorrectly assume the woman is not clapping if we view the frame where their hands are apart. We do this by creating an optional parameter called window size. The window size refers to the number of frames the Retriever extracts each time. When the window size is non-zero, the Retriever still specifies a single timestamp to go to, but when retrieving the frame at that timestamp, we also take the number of frames specified by the window size before and after the selected frame.
\section{Evaluation}
\label{sec:expr}

We evaluated our {\smodel} framework on several benchmarks described in~\Secref{sec:eval:datasets}, and compared it to multiple baselines in \Secref{sec:eval:baselines}. The results and ablations are in \Secref{sec:eval:results} and \Secref{sec:eval:ablations}. Additional results and ablations are in the Supplementary in~\Secref{supp:expr}.

\subsection{Datasets}
\label{sec:eval:datasets}

We use the following datasets: \textbf{(1) NExT-QA~\citep{xiao2021nextqanext}} is a dataset that tests causal action reasoning and temporal understanding. Following the trend of works before us, we evaluate our method on the 5,000 questions in the NExT-QA validation set. \textbf{(2) EgoSchema ~\citep{mangalam2023egoschema}} is a challenging dataset that tests long-form video understanding. Viewers need to view 100 seconds of the video on average to answer the question correctly. \textbf{(3) STAR~\citep{wu2021star_situated_reasoning}} tests reasoning in real-world video situations. \textbf{(4) Perception Test~\citep{pătrăucean2023perception}} is a challenging dataset that focuses on skills such as memory, abstraction, physics, and semantics and is intended to be approached in a few-shot or zero-shot manner. 


\subsection{Baselines}
\label{sec:eval:baselines}
In our experiments, we compare our method to recent state-of-the-art zero-shot (ZS) methods, such as LLoVi~\citep{zhang2023simple}, ProViQ~\citep{choudhury2023zeroshot}, and other methods that are not necessarily ZS, such as SeViLA \citep{yu2023self}, and MC-ViT \citep{balažević2024memory}. We note that SeViLA uses fine-tuned components on QV-Highlights~\citep{lei2021qvhighlights}, while MC-ViT is fine-tuned on NExT-QA for Perception Test. Additional baselines are in Supplementary in~\Secref{supp:expr:more_results}.

\subsection{Results}
\label{sec:eval:results}

Our results are shown in \tabref{table:results}. We use GPT-4 for NExT-QA to ensure a fair comparison with LLoVi, which is the current state-of-the-art that uses GPT-4 to uniformly caption frames across the entire video. Interestingly, our method outperforms LLoVi despite viewing 50\% fewer frames on average. Second, we also outperform SeViLA by +4.6\%, although SeViLA uses a keyframe selector that is fine-tuned on a video moment retrieval and grounding task while our method is fully ZS.

For EgoSchema, we use GPT-4 with a retrained version of LaViLa that excludes overlapping Ego4D and EgoSchema videos to prevent data leakage. We show strong performance on long-form videos, where we outperform LLoVi by +3.0\%, while viewing 95\% fewer clips on average.



We use GPT-3.5 for Perception Test and STAR because it is cheaper than GPT-4, but results are likely to be improved even further with GPT-4. Nevertheless, we achieve higher accuracy than LongViViT on Perception Test by +4.5\% and MC-ViT by +2.1\%, although it was fine-tuned on NExT-QA. On STAR, we surpass the best ZS approach by +2.7\% and the best fine-tuned result by +0.3\%. 


Finally, please refer to~\Secref{supp:expr} for more experimental and ablation results, and~\Secref{sec:supp:qual} for additional visualizations. For example, in~\figgref{fig:supp_nextqa_comparison}, we see that question-answering is able to extract more relevant details in comparison to simple captioning. This may explain why our method significantly outperforms the descriptive split of NExT-QA. In~\figgref{fig:perception_supp_success}, we see our method is able to reason about vague references and correctly identify relevant objects through question-answering.

\subsection{Agent Ablations}
\label{sec:eval:ablations}

We perform ablations using 1000 randomly selected questions from the NExT-QA training set (see \tabref{tab:agent_ablations}). Unless specified, we use GPT-3.5/LLaVA-1.6 as the LLM/LMM for all agents.



\minisection{Ablating the Planner}
The Planner module outputs a plan, a list of instructions that guides the behavior of all other modules. We test the impact of removing it from our framework and find that it is worse. We hypothesize this is because the Planner provides many temporal cues that guide the Retriever module's search, such as ``go to the middle of the video'', and without these cues, the Retriever is not as good at selecting the next timestamp. Moreover, the Planner also helps the Evaluator better decide when to stop since in our iterative approach, the Evaluator uses the plan to determine when to stop. We also try removing re-planning by only running the Planner once in the beginning and keeping this plan fixed throughout. We find that this also reduces performance, showing the importance of adjusting plans to new information.

\minisection{Ablating the Retriever} 
The Retriever module determines the next timestamps to view, which helps focus our information collection. We ablate it by uniformly sampling frames from the video at 2-second intervals, similar to other methods like LLoVi, which performs worse. We believe this is because the Retriever allows us to capture frames that might have otherwise been skipped through uniform sampling and selects fewer unimportant frames that might mislead the model.

\begin{table}[t!]
    \caption{Ablations on the agents in our framework.
    }
    \label{tab:agent_ablations}
    \centering
    \small
    \begin{tabular}{lccc}
        \toprule
        Agent & Ablation & Avg. (\%) \\
        \midrule
        -------- & Baseline & 60.4 \\ 
        \midrule
        \multirow{2}{*}{Planner} & Removal & 58.1 \\
        & Fixed Plan & 58.1 \\
        Retriever & Removal & 56.9 \\
        Extractor & Caption Only & 58.2 \\
        Evaluator & No Iteration & 56.8 \\
        Summarizer & Removal & 57.2 \\
        \bottomrule
    \end{tabular}
\end{table}

\minisection{Ablating the Extractor}
Question asking is important as it allows us to capture more fine-grained and question-relevant information compared to simple caption generation, which produces a generic description. Thus, we ablate the Extractor by only allowing the LMM to caption frames. We find that this decreases performance by -2.2\%, suggesting that the ability to ask specific questions about a frame is important. We notice many generated captions capture the main idea of visual information in the frame, but are lacking in fine-grained details.

\minisection{Ablating the Evaluator}
The Evaluator reflects on the collected information and decides if there is enough information to answer the question. To examine the impact of this reflection process, we make the Evaluator answer on the first iteration, finding it results in a -3.6\% performance decrease.

\minisection{Ablating the Summarizer} When collecting large amounts of information from videos, we use a Summarizer to condense the information, since long inputs can be challenging for LLMs. This has also been observed in recent work ~\citep{zhang2023simple}. To understand the impact of the Summarizer, we remove it. The results indicate that removing it degrades performance by -3.2\%, demonstrating the advantage of more concise information.

\renewcommand{\thefootnote}{\fnsymbol{footnote}}
\begin{table*}[tb!]
    
    \vspace{-0.2em}
    \caption{\textbf{Ablation Results.} We perform ablations on 1000 randomly selected questions from the NExT-QA training set. We report (a), (b) replacing different LLMs and LMMs, (c) selecting different numbers of frames to view in the Retriever, and (d) changing the number of questions asked in the Extractor. We use GPT-3.5/LLaVA-1.6 as the LLM/LMM, 5 frames for the Retriever, and 3 questions for the Extractor, unless otherwise specified.}
    \centering
    \begin{subtable}[t]{.3\linewidth}
	\tablestyle{5pt}{1.25}
        \centering
        \caption{{Replacing Diff. LLMs}}
        \label{tab:agents_llm}
            \begin{tabular}{llccc}
        \toprule
        LLM & LMM & Accuracy \\
        \midrule
        GPT-3.5 & \multirow{4}{*}{LLaVA-1.6} & 60.4 \\
        Llama 3 &  & 63.9 \\
        GPT-4 &  & 65.8 \\
        GPT-4o &  & 68.0 \\
        \bottomrule
    \end{tabular}
    \end{subtable}
    \begin{subtable}[t]{.25\linewidth}
	\tablestyle{5pt}{1.25}
        \centering
        \caption{{Replacing Diff. LMMs}}
        \label{tab:agents_lmm}
            \begin{tabular}{lccc}
        \toprule
        LLM & LMM & Accuracy \\
        \midrule
        GPT-4 & GPT-4V & 64.7 \\
        \hline
        \multirow{3}{*}{GPT-3.5} & BLIP-2 & 52.7 \\
        & GPT-4V & 59.5 \\
        & LLaVA-1.6 & 60.4 \\
        \bottomrule
    \end{tabular}
    \end{subtable}
    \hspace{18pt}
    \begin{subtable}[t]{0.15\linewidth}
        \tablestyle{3pt}{1.25}
        \centering
	    \caption{{\# of Frames}}
        \label{tab:retriever}
	    \begin{tabular}{cc}
        \toprule
        \# Frames & Accuracy\\
        \midrule
        1 & 59.0 \\
        3 & 57.9 \\
        5 & 60.4 \\
        7 & 59.0 \\
        \bottomrule
    \end{tabular}
    \end{subtable}
    \hspace{8pt}
    \begin{subtable}[t]{0.2\linewidth}
	\tablestyle{3pt}{1.25}
        \centering
        \caption{{\# of Questions}}
        \label{tab:questions}
            \begin{tabular}{cc}
        \toprule
        \# Questions & Accuracy\\
        \midrule
        0 & 58.1 \\
        1 & 58.4 \\
        3 & 60.4 \\
        5 & 59.6 \\
        \bottomrule
    \end{tabular}
    \end{subtable}
    \label{tab:ablations}
\end{table*}

\renewcommand{\thefootnote}{\arabic{footnote}}

\subsection{Additional Experiments}
\label{sec:eval:add_expers}

\minisection{Substituting different LLMs/LMMs}
To see how the choice of the LLM and LMM affects our framework's performance, we swap different LLMs and LMMs into our framework (see \tabref{tab:agents_llm} and \tabref{tab:agents_lmm}). 
We first try different LLMs while fixing the LMM to be LLaVA-1.6. While GPT-4 performs better than GPT-3.5 by a significant margin of +5.4\%, open-source model Llama 3 is very close (-1.9\%), while incurring no additional cost. We also evaluate the newly released GPT-4o, which outperforms GPT-4 by +2.2\% while being 61\% cheaper, showing that our method can leverage better future models. Second, we use different LMMs while fixing the LLM to be GPT-3.5. We find that LLaVA-1.6 does best, GPT-4V is slightly worse (-0.9\%), and BLIP-2 is significantly worse (-7.7\%). Finally, we run an experiment using GPT-4V as both the LLM and LMM, and find that this does worse than GPT-4 and LLaVA-1.6 by -1.1\%.

\minisection{GPT-4V Baseline} To get a baseline using GPT-4V, we use a method similar to LLoVi using a subset of 500 examples. We use GPT-4V to caption the video uniformly, then ask it to answer the question given the choices, which results in a performance change of -2.0\%. We note our motivation is to refrain from captioning every single frame, instead finding frames that help us answer the question.

\minisection{Retriever window size}
We experiment with different window sizes in~\tabref{tab:retriever} which is the number of frames the Retriever extracts centered around the selected frame. This allows us to capture better actions that occur quickly or require more context to understand. We find that choosing 5 frames yields the best results and a +1.4\% increase when compared to selecting a single frame, but viewing more than 5 decreases performance. This suggests retrieving multiple frames can help the model better capture relevant information, but retrieving too many frames can lead to too much information.


\begin{figure}[ht]
    \centering
    \includegraphics[width=\columnwidth]{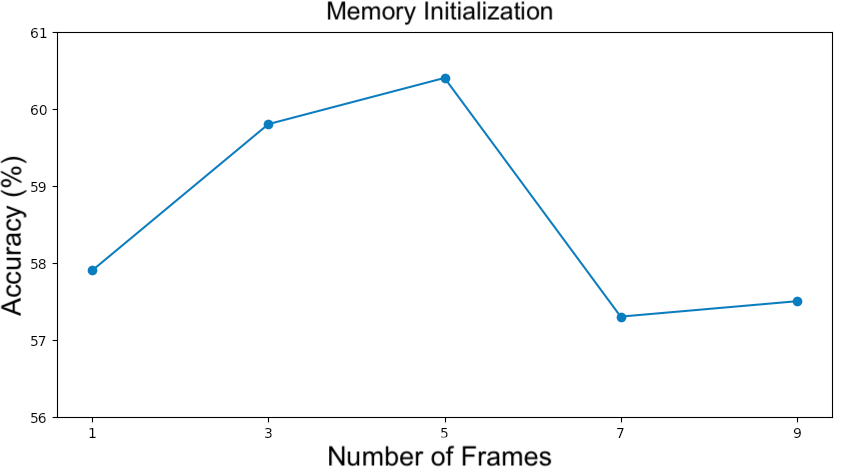}
    \captionsetup{width=0.9\columnwidth}
    \captionof{figure}{Comparison of different Memory Initialization (1, 3, 5 frames). 5 frames is optimal.}
    \label{mem_init}
\end{figure}

\begin{table*}[ht]
\centering
\caption{Comparison of TraveLER with other few-shot and zero-shot keyframe localization methods. For fair comparisons, we \textcolor{gray}{gray out} methods with fine-tuned components in their model. The best scores are in \textbf{bold}.}
\begin{tabular}{lcccc}
\toprule
Method & \multicolumn{4}{c}{NExT-QA (Random Subset)} \\
\cmidrule{2-5}
& Temporal & Causal & Descriptive & Average \\
\midrule
\rowcolor{gray!20} SeViLA - Localizer \citep{yu2023self} & 48.8 & 61.2 & 68.3 & 58.2 \\
\rowcolor{gray!20} Moment-DETR \citep{lei2021qvhighlights} & 45.3 & 55.8 & 70.8 & 54.6 \\
\midrule
SigLIP \citep{zhai2023sigmoid} & 48.4 & 61.5 & \textbf{73.8} & 59.1 \\
TraveLER - Planner \& Retriever (ours) & \textbf{50.9} & \textbf{62.7} & 72.4 & \textbf{60.3} \\
\bottomrule
\label{tables:keyframe_comparison}
\end{tabular}
\end{table*} 

\minisection{Memory Intialization}
In order for the Planner to create effective plans, it is beneficial to initialize the memory bank properly. Memory initialization allows the Planner to have a high-level overview of the video, and create a corresponding plan on how to traverse the video given the initial frames. We perform three different initializations with 1, 3, 5, 7, and 9 frames and display our results in ~\figgref{mem_init}. We observe that initializing the memory bank with 5 frames uniformly sampled from the video (0, 0.25, 0.5, 0.75, 1 for beginning, quarter, middle, three-quarter, end) yields the best result. In contrast, we notice a decrease in accuracy of -2.9\% when using 9 frames, -3.1\% when using 7 frames, -0.6\% when initializing with 3 frames (0, 0.5, 1 for beginning, middle, end), and -2.5\% when initializing with 1 frame (0.5 for middle).

\minisection{Robustness to choosing incorrect frames} To see if our approach can recover from viewing incorrect frames, we choose random frames for the first 3 iterations (out of 5 total iterations) before using the Retriever. This leads to an accuracy drop of only 2.8\%, showing our method is capable of recovering from viewing incorrect frames.


\minisection{The number of questions}
Question answering allows us to extract more specific details from our visual inputs. However, we noticed that too many questions can yield irrelevant questions and false positives. Thus, we experiment with modifying the number of questions asked for each frame by our Extractor (see~\tabref{tab:questions}). We record results for a 5-question, 3-question, and 1-question maximum. Note that 0-questions asked is equivalent to only allowing captions, which is discussed in the Extractor ablation. From our results, we find that a 3-question limit yields the best results compared to asking 1 or 5 questions (+2.0/+0.8\%). This suggests that asking questions helps in extracting relevant information, but too many questions can lead to false positives or too much irrelevant information.

\minisection{Memory bank initialization and formatting}
Our memory bank $M$ stores information that all modules rely on to make decisions. First, we experiment with different initializations as $M$ must be initialized in the first iteration. We experiment with initialization of 1, 3, and 5 uniformly sampled captions. We find using 5 evenly spaced frames yields the best results, possibly because it starts the model with a general overview of the video before it starts to collect more relevant information. We also experiment with changing the format of the memory bank from JSON to markdown tables. We find this does worse by -2.9\%, possibly because LLMs are better able to understand JSON formats.



\minisection{Comparison with other keyframe selection methods}
In~\tabref{tables:keyframe_comparison}, we compare our Planner and Retriever with other keyframe localization methods by replacing our Planner and Retriever with each of the other methods, and use our Extractor and Evaluator to perform the question answering. For all these methods, we use GPT-3.5 and LLaVA-1.6, and we evaluate these methods on a random subset of 1000 examples from the training set of NExT-QA. Note that other methods find keyframes in one inference iteration, whereas our inference occurs over multiple iterations. Therefore, to ensure fair comparisons, we uniformly sample 32 frames and extract out 4 keyframes in the other methods, and we run 4 iterations of {\smodel} using the Retriever with a window size of 2 to similarly find 4 keyframes among fewer than 32 viewed frames.

We find that our Planner and Retriever surpasses other keyframe localization methods, despite considering fewer total frames ($\sim$ 25 total frames; we have 5 for memory initialization, and up to 5 frames each iteration). We would like to highlight that while our method is effective at finding keyframes, we do not need to find all keyframes to answer a question. Instead, we are often able to choose the correct answer with only a subset of the keyframes.

\minisection{Cost of inference} We test costs for GPT-3.5 and GPT-4, which have costs per inference of \$0.03 and \$0.67, respectively. We also test the open-source model Llama 3, which is free to run with only a minor performance decrease (-1.9\%). Finally, we also try the new GPT-4o model, which is both better (+2.2\%) and cheaper (61\%) than GPT-4. We believe this trend will only continue as models become better and cheaper in the future.

\section{Conclusion}
\label{sec:discussion}


We design a modular, multi-LMM agent framework for video-question answering based on several agents with different roles, instructed by a Planner agent that updates its instructions using shared feedback between the other agents. Our method creates a plan to ``traverse'' through the video, asking questions about individual frames to ``locate'' and store key information, and then ``evaluate'' if there is enough information to answer the question, ``replanning'' using new feedback if necessary. Through extensive experiments and ablations, we find that the proposed \textit{TraveLER} approach is not only easy to employ with different models but also improves performance on several video question-answering benchmarks without the need to fine-tune on specific datasets.

\clearpage
\clearpage
\section{Limitations}
\label{sec:limitations}
In this work, we present a modular, zero-shot framework for video question answering (VideoQA) and demonstrate its effectiveness by improving on multiple state-of-the-art benchmarks. While {\smodel} offers substantial benefits for VideoQA, it is important to recognize certain limitations that accompany our approach. Firstly, the effectiveness of our model relies heavily on the strength of the LLM and LMM. We notice that false-positives and incorrect statements from the LMM heavily impact performance. Our method's runtime also depends on the runtime of current existing methods, and with the modularity of our method we expect this to improve with faster and better models in the future. 
Finally, we do not anticipate negative impacts of this work, but, as with any Machine Learning method, we recommend exercising caution. 


\section{Acknowledgements}
We would like to thank Suzie Petryk, Chancharik Mitra, Alon Mendelson, David Chan, Assaf Arbelle, and Leonid Karlinsky for helpful feedback and discussions. This project was supported in part by DoD, including PTG and/or LwLL programs, as well as BAIR's industrial alliance programs.

\bibliography{custom}

\appendix

\appendix
\clearpage

\section*{Supplementary Material for ``TraveLER''}

Here we provide additional information about our experimental results, qualitative examples, implementation details, and datasets. Specifically, \Secref{supp:expr} provides more experiment results, \Secref{supp:impl} provides additional implementation details, \Secref{supp:rw} provides additional related work, and \Secref{sec:supp:qual} provides qualitative visualizations to illustrate our approach.

\section{Additional Experiment Results}
\label{supp:expr}
We begin by presenting several additional ablations in~\Secref{supp:more_ablt} that further demonstrate the benefits of our \smodel~approach. We also present additional results in~\Secref{supp:expr:more_results}.

\subsection{Additional Ablations}
\label{supp:more_ablt}

In what follows, we provide additional ablations that further illustrate the benefits of {\smodel}. For all ablations, we compare the ablated experiment with the corresponding best-performing {\smodel} results on a random sample of 1000 examples from the training set of the NExT-QA dataset. We use GPT-3.5 as the LLM and LLaVA-1.6 as the LMM.

\begin{figure}[t!]
    \centering
    \includegraphics[width=\columnwidth]{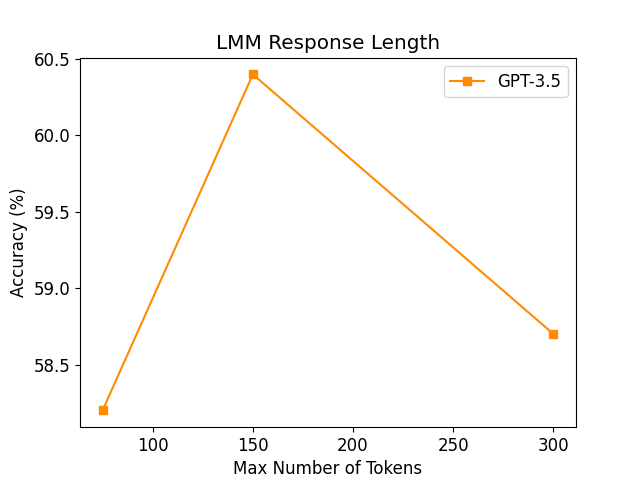}
    \captionsetup{width=0.9\columnwidth}
    \captionof{figure}{Comparison of different LMM Response Length (75, 150, 300 max tokens). 150 is optimal.}
    \label{lmm_response_length}
\end{figure}

\minisection{LMM response length}
The LMM in our framework is crucial because it allows us to capture more relevant and question-specific details from visual input. However, if the LMM's responses are too long, the memory bank will become too large, whereas if the LMM's responses are too short, insufficient information will be captured. Thus, we conduct an experiment to determine the optimal LMM response length, and display our results in Fig \ref{lmm_response_length}. We find that limiting the LMM response to 150 tokens yields the most optimal performance, while accuracy decreases by -2.2\% and -1.7\% if the response is limited to 75 tokens and 300 tokens respectively. This supports the fact that there is a tradeoff between not collecting enough information for short response lengths and collecting too much information as the LMM response size increases.

\minisection{Prompt analysis}
In each module of our framework, we use a task prompt to provide instructions to our agents (LLMs or LMMs). The construction of these prompts plays a large role in how instructions are executed. Currently, we use the question $Q$ as input into all prompts $(P_\mathrm{T}^{(1)}), (P_\mathrm{R}^{(1)}), (P_\mathrm{E}^{(1)}), (P_\mathrm{V}^{(1)})$. However, we use the choices $C$ as input only for the Planner and Evaluator prompts since the Planner needs the choices to tailor its plan, and the Evaluator needs the choices to answer the question. We experiment with adding the choices $C$ to the Retriever and Extractor prompts, and find that this degrades performance by -1.6\%. This may be because the incorrect choices mislead the Retriever into searching for non-existent events or the Extractor into asking irrelevant questions.

\subsection{Additional Results}
\label{supp:expr:more_results}

\minisection{Results on Causal-VidQA} We also report zero-shot results on Causal-VidQA \citep{li2022from}, a dataset designed to focus on causal related questions to facilitate deeper video understanding towards video reasoning. We see that our method outperforms the best zero-shot method by 16.9\%.

\minisection{Additional Baselines}
We also compare our approach with the concurrent work MoReVQA~\citep{min2024morevqa}. It can be seen that MoReVQA performs better on NExT-QA by 1\%, while our method performs better on the more difficult long-form video understanding EgoSchema benchmark by 1.6\%. We also note some key differences. Firstly, our method is able to re-plan based on feedback from previous iterations while MoReVQA uses a single forward pass through the stages and cannot modify its approach during the process. Secondly, we use an iterative approach while MoReVQA does not, which allows our method to adjust and improve using feedback from previous iterations.


\begin{table}[t!]
    \caption{Zero-shot (ZS) results on Causal-VidQA.
    }
    \centering
    \begin{tabular}{lcc}
        \toprule
        Model & Acc. (\%) \\
        \midrule
        Just-Ask \citep{yang2021justask} & 27.1 \\
        CaKE + CoMem \citep{su2023language} & 30.0 \\
        CaKE + HGA \citep{su2023language} & 30.6 \\    
        \rowcolor{lavendermist}
        TraveLER (ours) & \textbf{47.5}\gcol{+16.9} \\
        \bottomrule
    \end{tabular}
\end{table}

\section{Additional Implementation Details}
\label{supp:impl}

To run our models on larger benchmarks, we use 8 NVIDIA RTX 6000 GPUs and split the dataset across multiple processes. In addition, we use the SGLang \citep{zheng2023efficiently} package, which provides a variety of performance optimizations for our LMMs and enables us to perform batched inference for models that do not natively support doing so. We serve our LMM on a single GPU and implement a queue that is shared across all runs. This allows individual runs to asynchronously call the LMM using an API request instead of creating a new instance of the LMM for each run. Typically, we have 4-5 processes sharing the same LMM. 

Next, we use all default parameters. For our LMMs, we experiment with modifying the maximum output token length as an ablation, but use all default parameters otherwise. For the Llama 3 ablation in \ref{tab:agents_llm}, we serve Llama 3 70B on 4 NVIDIA A100s using vLLM \citep{kwon2023efficient}. We report results from a single run for all experiments.

\subsection{Prompts}
\label{supp:impl:prompts}

Our prompts are shown in \tabref{tab:prompt_plan}, \tabref{tab:prompt_retr}, \tabref{tab:prompt_extract}, and \ref{tab:prompt_eval}. The black text is the base prompt template, and we replace the blue text with the corresponding information from the relevant video. The generated outputs are in the orange text.

\begin{table*}[t!]

\centering
\begin{minipage}{1.99\columnwidth}\vspace{20mm}    
    \centering
    \begin{tcolorbox}[width=0.9\textwidth]
    \textbf{User:} \\
    Create the best plan to gather information to answer the question.

    QUESTION: \textcolor{blue}{QUESTION}
    CHOICES: \textcolor{blue}{CHOICES}

    You are provided information collected from individual frames of the video and is represented as a dictionary keyed by the timestamps of the frames. \\
    INFORMATION: \textcolor{blue}{INFO}

    You are also given an explanation for why you aren't able to definitively answer the question with the current information. \\
    EXPLANATION: \textcolor{blue}{EXPLANATION}

    Follow these rules:\\
    1. You only have access to individual frames of the video, with no audio. You can go to a certain timestamp, search for actions or settings, and describe or ask questions about individual frames. \\
    2. Make sure that you have viewed the relevant frames.

    Make your plan as simple and straightforward as possible, and no longer than 5 steps long. Return your plan as a numbered list, after PLAN. Do not include any other response or explanation. Let's think step-by-step.
    
    \rule[0.25\baselineskip]{\textwidth}{1pt}
    \textbf{Assistant:}
    \textcolor{orange}{Output: PLAN}
\end{tcolorbox}
\caption{Planner prompt $P_T$:}
\label{tab:prompt_plan}
\end{minipage}
\end{table*}

\begin{table*}[t!]
\centering
\begin{minipage}{1.99\columnwidth}\vspace{20mm}    
    \centering
    \begin{tcolorbox}
    \textbf{User:} \\
    You are given the following information about a \textcolor{blue}{LENGTH} second video, with information from individual frames at different timestamps. \\
    INFORMATION: \textcolor{blue}{INFO}

    PLAN: \textcolor{blue}{PLAN}

    Currently, you are viewing second \textcolor{blue}{CURR}. Choose the timestamp, in seconds, of the next frame to view. When choosing the next frame to view, remember that you are trying to collect information to answer this multiple choice question: \textcolor{blue}{QUESTION}

    Think of what information you need, and consider what information you already have. Use the temporal nature of the video and your past information to choose the next frame. Do not choose a frame you already have information about, and make sure that the frame you choose is at least \textcolor{blue}{WINDOW SIZE} seconds apart from the second you are currently viewing.

    Return your answer as a single Python float representing the second you want to view. Don't provide any other response or explanation.
    
    \rule[0.25\baselineskip]{\textwidth}{1pt}
    \textbf{Assistant:}
    \textcolor{orange}{Output: TIMESTAMP}
    \end{tcolorbox}
   
\end{minipage}
\caption{Retriever prompt $P_R$:}
\label{tab:prompt_retr}
\end{table*}
\begin{table*}[t!]

    \centering
    \begin{minipage}{1.99\columnwidth}\vspace{20mm}    
        \centering
        \begin{tcolorbox}
        \textbf{User:} \\
        You are given the following information about a \textcolor{blue}{LENGTH} second video, with information from individual     frames at different timestamps. \\
        INFORMATION: \textcolor{blue}{INFO}
    
        Currently, you are viewing second \textcolor{blue}{CURRENT TIMESTAMP}, which has the caption: \textcolor{blue}{FRAME CAPTION}
        
        Form up to 3 questions about this frame to best help answer the multiple-choice question: \textcolor{blue}{QUESTION}.
    
        Follow these rules: \\
        1. Use the given information to decide what further visual information you need to answer the question. \\
        2. Since you are asking questions about a single frame, you cannot ask about other frames, reference past or future events,     or ask about specific timestamps.
    
        Return your questions as a Python list of strings (in double quotes) and don't include any numbered lists, backticks, or    language hints. Follow Python syntax. Make sure you have followed the steps. Don't provide any other response or explanation.
        \rule[0.25\baselineskip]{\textwidth}{1pt}
        \textbf{Assistant:}
        \textcolor{orange}{Output: QUESTIONS}
    \end{tcolorbox}
    \end{minipage}
    \caption{Extractor prompt $P_E$:}
    \label{tab:prompt_extract}
\end{table*}
\begin{table*}[t!]

\centering
\begin{minipage}{1.99\columnwidth}\vspace{20mm}    
    \centering
    \begin{tcolorbox}
        \textbf{User:} \\
        Evaluate if there is enough information to answer a multiple-choice question about a    video and if the plan has been completed.
    
        If there is enough information to choose the correct answer with complete certainty     and the plan has been followed, return the index of the choice after a brief    explanation. Otherwise, return None after a brief explanation of why you can't narrow  down to a single answer choice. Be strict and don't guess.
    
        INFORMATION: \textcolor{blue}{INFO}\\
        PLAN: \textcolor{blue}{PLAN}\\
        QUESTION: \textcolor{blue}{QUESTION}\\
        CHOICES: \textcolor{blue}{CHOICES}
    
        Give a brief explanation. Then, include your final answer after the words "Final    Answer:" in your response at the end. Do not include anything other than the answer as an integer or None after "Final Answer:".
    
        Let's think step by step.
        
        \rule[0.25\baselineskip]{\textwidth}{1pt}
        \textbf{Assistant:}
        \textcolor{orange}{Output: ANSWER}
        
    \end{tcolorbox}
\end{minipage}
\caption{Evaluator prompt $P_V$:}
\label{tab:prompt_eval}
\end{table*}

\subsection{NExT-QA}
\minisection{Dataset}
NExT-QA is a challenging dataset that tests causal action reasoning and temporal understanding. It contains 5,440 videos with an average length of 44s. Compared to earlier VideoQA benchmarks ~\citep{papalampidi2023simple, xue2017unifying, zeng2016leveraging, zhu2017uncovering}, NExT-QA requires going beyond simple recognition of objects and actions to answer the questions correctly. Each question requires selecting the best option out of 5 choices, often with very similar degrees of plausibility. Additionally, each question is categorized into either a Temporal, Causal, or Descriptive type. Temporal questions often ask what happens during, before, or after an event or action, while causal questions require advanced reasoning and inference about why an event or action occurs. Following the trend of works before us, we evaluate our method on the 5,000 questions in the NExT-QA validation set, which consist of 500 different videos. This dataset is in English.

\minisection{Inference Details} 
For NExT-QA, we use LLaVA-1.6 (Vicuna 13B) as the LMM and GPT-4 (gpt-4-1106-preview) as the LLM. We used the longer, comprehensive prompts, with no answer choices included in the Extractor prompt. We also initialize the memory bank with 5 frames, and use the multi-frame Retriever with 5 frames.
\label{supp:qual_vis}

\subsection{Perception Test}

\label{supp:impl:perception}
\minisection{Dataset}
In comparison with earlier VideoQA datasets \citep{maharaj2017dataset, zhu2017uncovering, zeng2016leveraging} that focus on computational tasks such as classification, detection, or tracking, Perception Test is a dataset that focuses on skills such as memory, abstraction, physics, and semantics. Moreover, it is designed to test the transfer capabilities of different models and intended to be approached in a few-shot or zero-shot manner. The dataset consists of 11.6k real-world videos with an average length of 23 seconds, and 38K multiple choice QA questions. This dataset is in English.

\minisection{Inference Details}
For Perception Test, we use LLaVA-1.6 (Vicuna 13B) as the LMM and GPT-3.5 (gpt-3.5-turbo-0125) as the LLM. We used shorter, simplified prompts, with no answer choices included in the Extractor prompt. We also initialize the memory bank with 5 frames, and use the multi-frame Retriever with 5 frames.

\subsection{EgoSchema}
\label{supp:impl:EgoSchema}
\minisection{Dataset} EgoSchema is a dataset that tests long-form video reasoning and is intended to be answered in a few-shot or zero-shot manner. It introduces the idea of certificate lengths, which are the minimum number of seconds it takes to be able to answer the question correctly. It has 5k questions in the full set, and a 500 question subset is used as the validation set. We use the full test set for evaluation. This dataset is in English.

\minisection{Inference Details}
For EgoSchema, we use GPT-4 (gpt-4-1106-preview) as the LLM and LaViLa as the LMM. To prevent data leakage, we use a retrained version of LaViLa that does not use overlapping videos between Ego4D and EgoSchema. For window size in the Retriever, we take the selected frame, along with the frame 1 second before and 1-2 seconds after the selected frame. 

\subsection{STAR}
\label{supp:impl:STAR}
\minisection{Dataset} STAR is a dataset that tests reasoning in real-world video situations. It consists of 22K video clips, with 60K situated reasoning questions, with 4 possible choices each. Questions are broadly divided into 4 main categories: interaction, sequence, prediction, and feasibility. This dataset is in English.

\minisection{Inference Details}
For STAR, we use LLaVA-1.6 (Vicuna 13B) as the LMM and GPT-3.5 (gpt-3.5-turbo-0125) as the LLM. We used shorter, simplified prompts, with no answer choices included in the Extractor prompt. We also initialize the memory bank with 5 frames, and use the multi-frame Retriever with 3 frames, since the videos are shorter.

\section{Additional Related Work}
\label{supp:rw}

\minisection{Modular Vision Frameworks} There has been a long history of work~\citep{andreas2017neural,herzig2018mapping,referential_relationships,baradel2018object,battaglia2018relational,herzig2019canonical,herzig2022orvit,herzig2019stag,herzig2023incorporating,2020ActionGraphs,avraham2022svit,Jerbi2020LearningOD,Herzig2022PromptonomyViT,MitraCCoT} that attempts to combine deep neural networks with modularity. Recently, works like VisProg~\citep{Gupta_2023_CVPR}, CodeVQA \citep{subramanian-etal-2023-modular}, RVP~\citep{ge2023recursive}, and ViperGPT \citep{surismenon2023vipergpt} have leveraged the improved coding capabilities of LMMs to generate code to compose different submodules together to answer visual questions. In addition, ProViQ \citep{choudhury2023zeroshot} extends ViperGPT's work in the video domain by adding more modules for VideoQA. Similarly, we leverage the strong power of LMMs in a modular approach. However, while these approaches have shown promising results, they are limited to single-shot planning when generating code, resulting in a fixed plan that cannot adapt. In contrast to these works, our approach has the advantage of being able to iteratively replan based on new information collected.

\section{Qualitative Visualizations}
\label{sec:supp:qual}

\begin{figure*}[tb!]
  \centering
  \includegraphics[width=\linewidth]{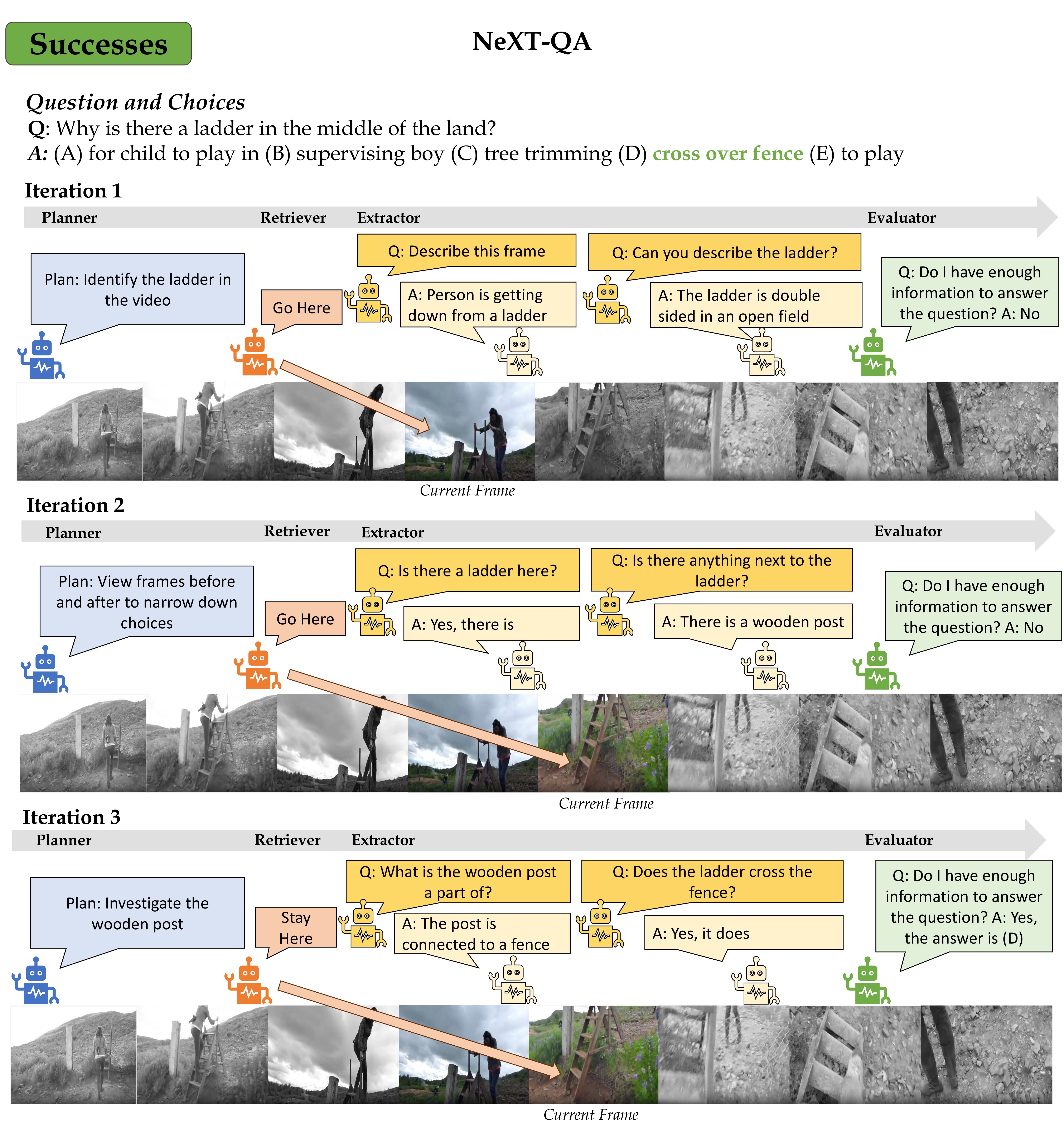}
  \caption{\textbf{NeXT-QA Success Predictions.} We can see that our framework can adapt to new information collected in past iterations. For example, in Iteration 3, our Planner module is able to use information about the wooden post from a previous iteration and ask further questions to identify the correct answer.}
    \label{fig:nextqa_supp_success}
\end{figure*}

\begin{figure*}[tb!]
  \centering
  \includegraphics[width=\linewidth]{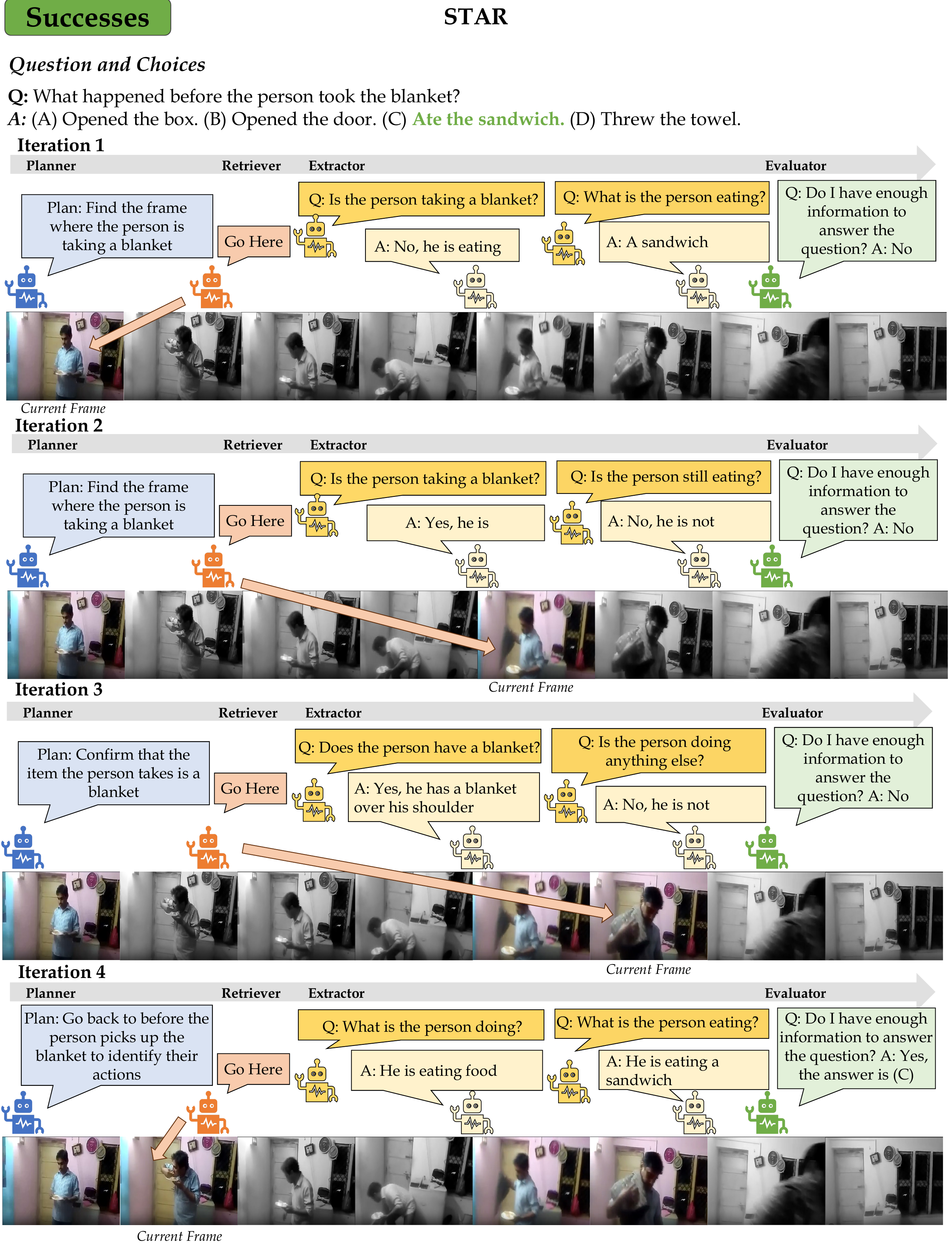}
  \caption{\textbf{STAR Success Predictions.} Here, we can see that our method does not require viewing frames sequentially. For example, we view the beginning of the video in Iteration 1, the middle of the video in Iterations 2 and 3, and return to the beginning in Iteration 4. Moreover, our method can collect information and double-check ambiguous information across different timestamps. For example, in Iteration 2, we are told the man is taking a blanket, and then we can view a different frame to confirm that he is indeed holding a towel in Iteration 3.}
    \label{fig:star_supp_success}
\end{figure*}

\begin{figure*}[tb!]
  \centering
  \includegraphics[width=\linewidth]{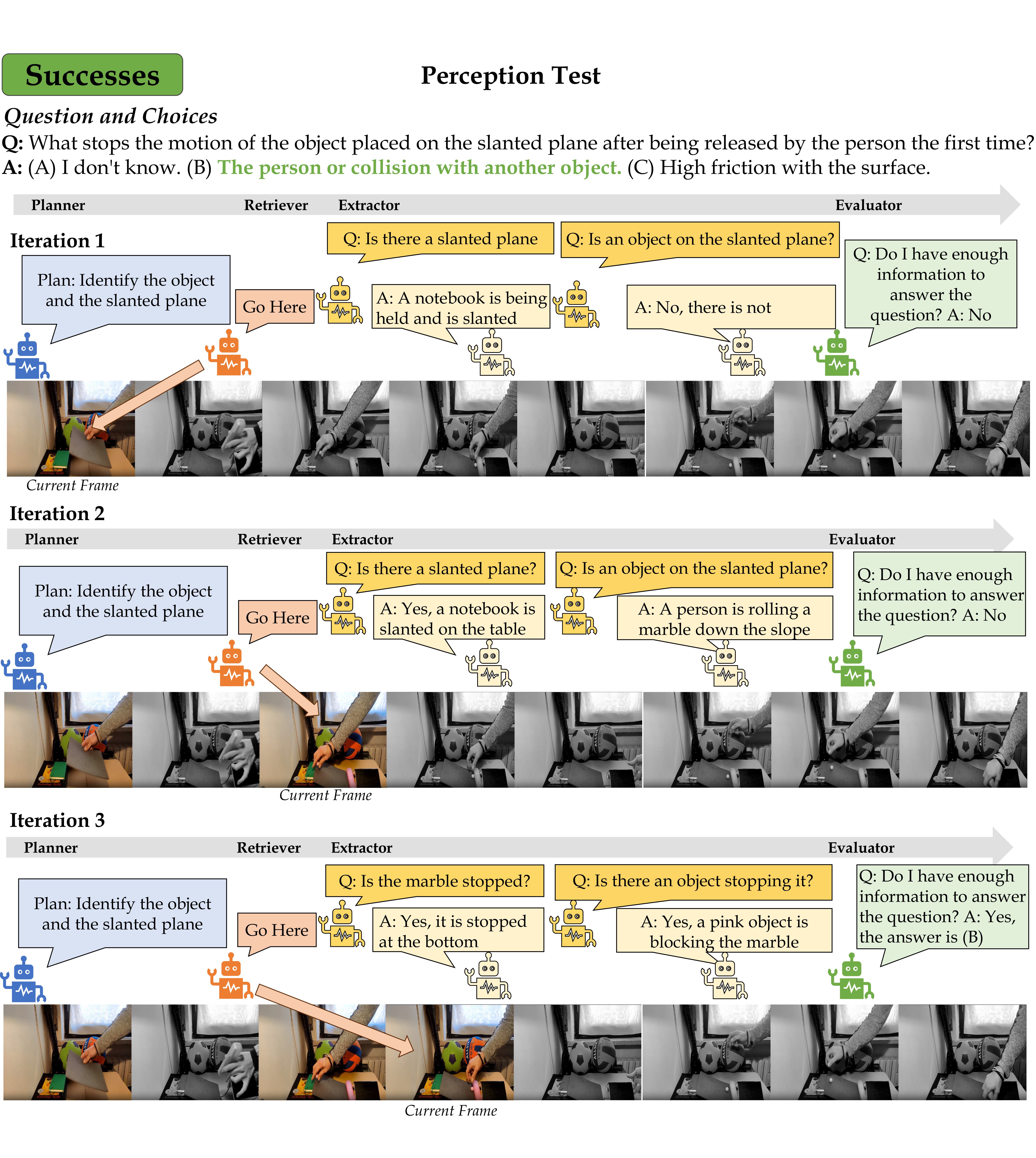}
  \caption{\textbf{Perception Test Success Predictions.} We display some success cases for the challenging Perception Test dataset. Here, our method is able to infer which objects the question refers to through our question-asking approach, even though the question does not explicitly describe them.}
    \label{fig:perception_supp_success}
\end{figure*}

\begin{figure*}[tb!]
  \centering
  \includegraphics[width=\linewidth]{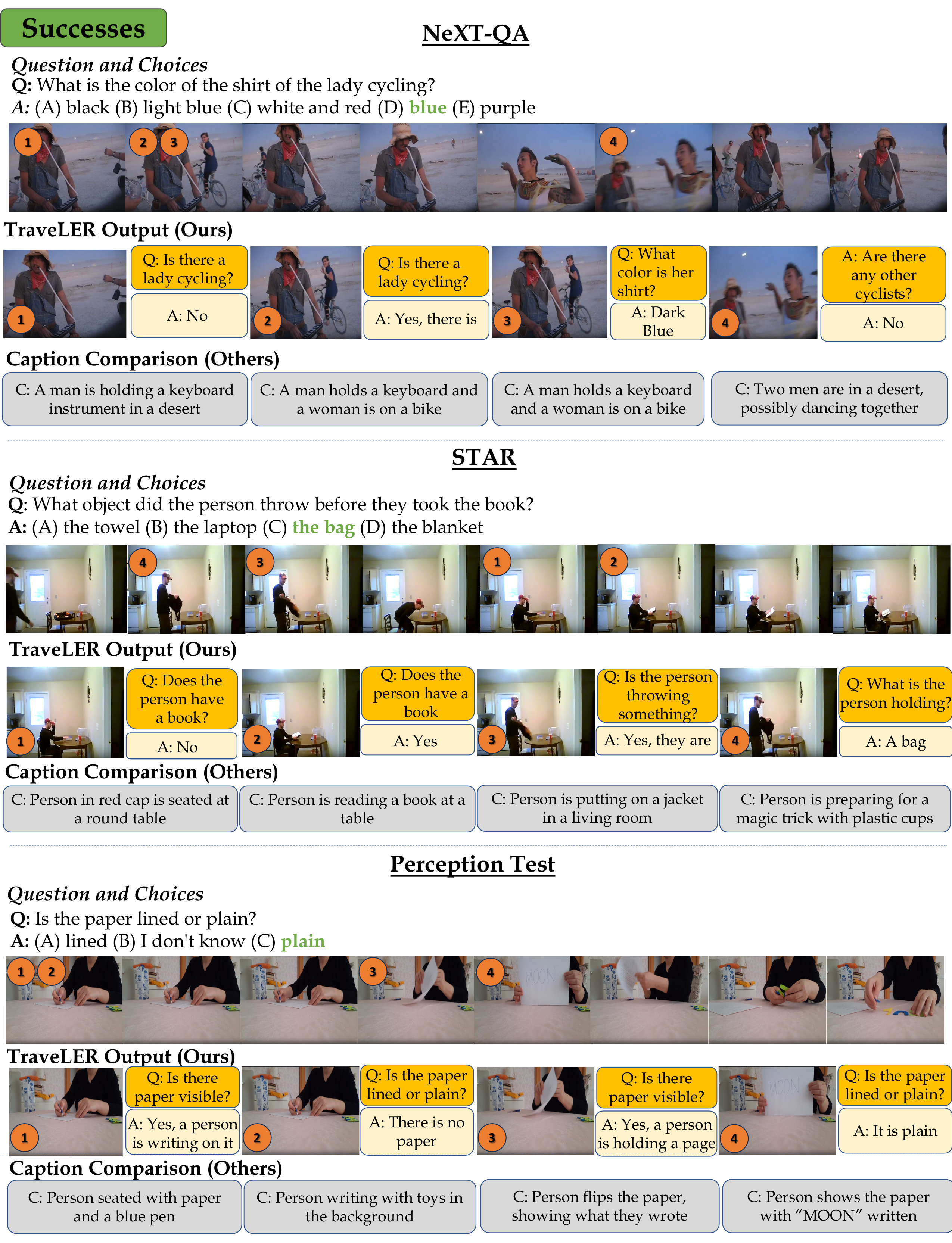}
  \caption{\textbf{Comparison with Captioning Approaches.} For each example, we display the videos on top, and the traversal order using the numbered orange circles. In the rows beneath, we display the frames in the order they are selected, and display corresponding Extractor question-answer output in yellow and captions in gray. We display compressed versions of GPT-4V-generated captions for a visual comparison. By asking specific questions, we can extract more detailed and relevant information than a general description.}
    \label{fig:supp_nextqa_comparison}
\end{figure*}

\begin{figure*}[tb!]
  \centering
  \includegraphics[width=\linewidth]{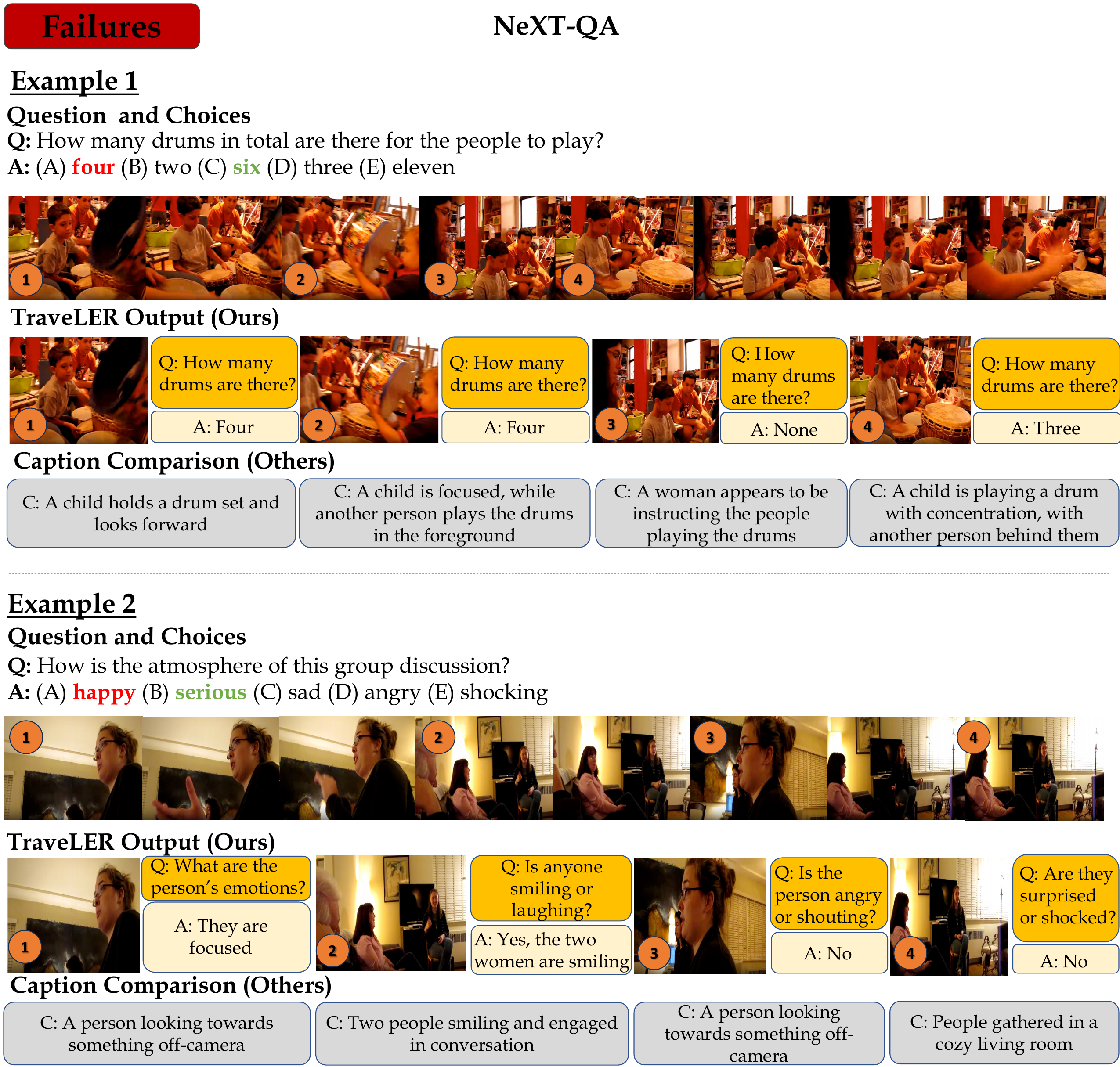}
  \caption{\textbf{NeXT-QA Failure Predictions.} Here, we display some failure cases for NeXT-QA. Like before, we display the video on top, and the traversal order using the numbered orange circles. In the row beneath, we display the frames in the order they are selected, and display corresponding Extractor question-answer output in yellow and captions in gray. We can see that conflicting information or false positives can mislead our approach. We also observe that counting can be a challenge for certain LMMs, but this can be mitigated in the future by swapping in stronger LMMs.}
    \label{fig:nextqa_supp_fail}
\end{figure*}

\begin{figure*}[tb!]
  \centering
  \includegraphics[width=\linewidth]{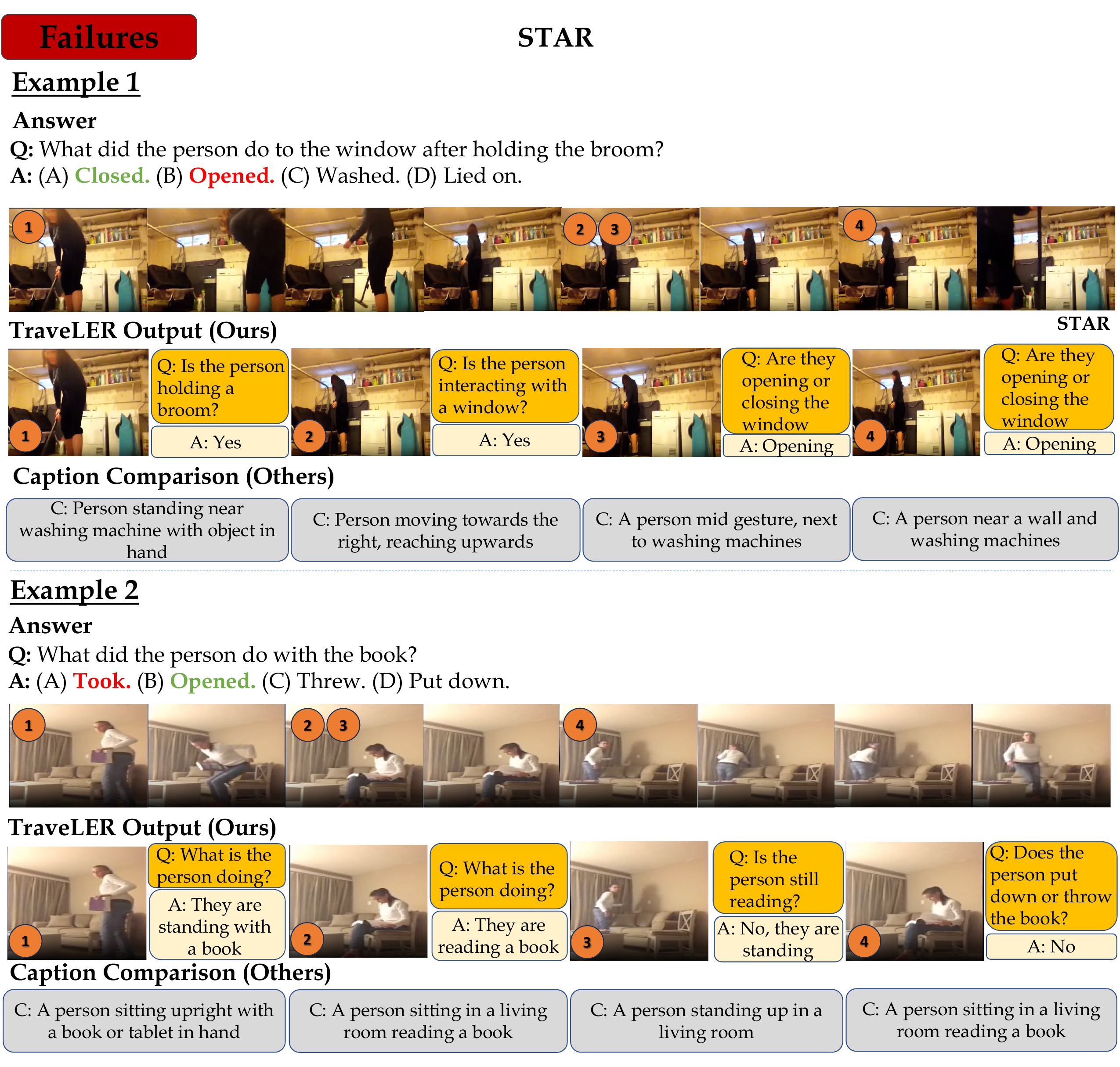}
  \caption{\textbf{STAR Failure Cases.} Here, we display some failure cases for the STAR dataset, using the same abridged representation described previously. We see that a limitation of a framewise approach is that it may be difficult to capture very temporal actions. For example, in Example 1, it is difficult to understand if the woman is opening or closing the window.}
    \label{fig:star_supp_fail}
\end{figure*}

\begin{figure*}[tb!]
  \centering
  \includegraphics[width=\linewidth]{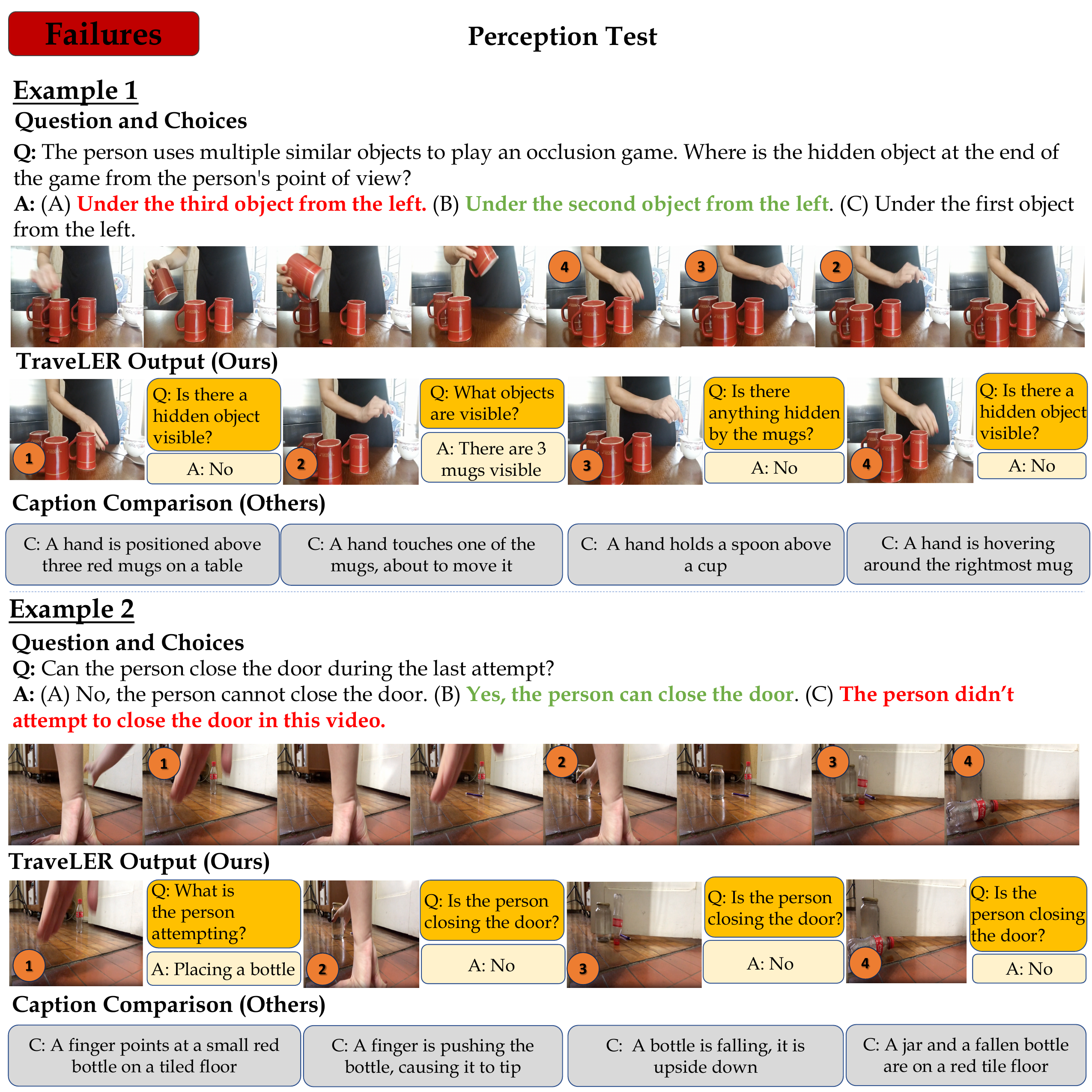}
  \caption{\textbf{Perception Test Failure Cases.} We display some qualitative visualizations for Perception Test failure cases  using the abridged representation discussed previously. We see that for some cases where objects of interest are occluded or not in the frame, our method might have difficulties extracting useful information.}
    \label{fig:perception_supp_fail}
\end{figure*}

\begin{figure*}[tb!]
  \centering
  \includegraphics[width=\linewidth]{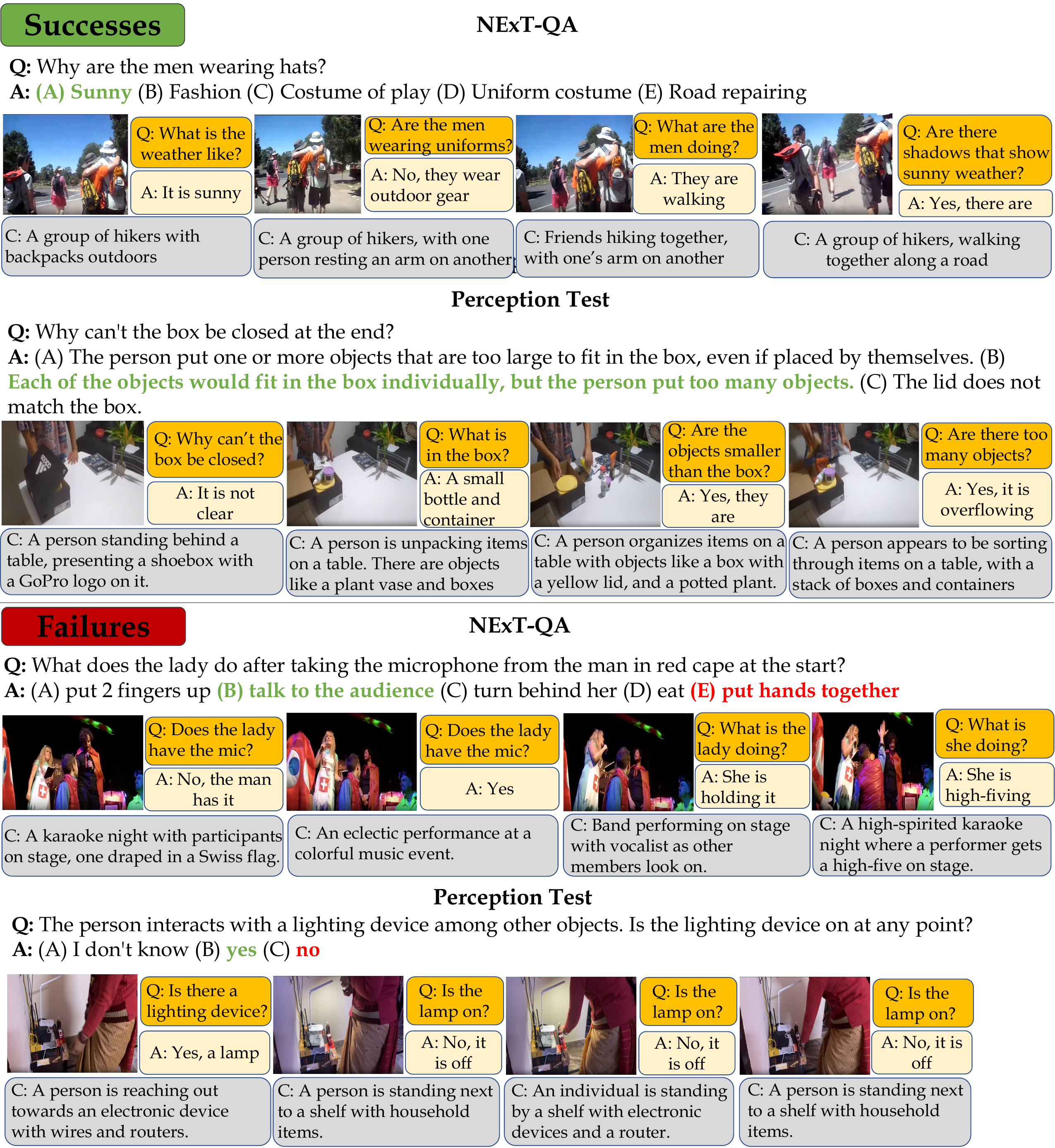}
  \caption{\textbf{An additional visualization of predictions.} We show more qualitative visualizations of our method on NExT-QA and Perception Test using our abridged representation, with successes on top and failures on the bottom. We compare our generated question-answer pairs for each frame (in yellow) with captions (labeled C in gray) generated from the same frame. We see that our method is able to extract more fine-grained and relevant information compared to simple captioning.}
    \label{fig:vis_results}
\end{figure*}

We present further qualitative success and failure cases of our {\smodel} framework. For each dataset, we display qualitative visualizations for 2 successes and 2 failure cases. For the success cases, we show expanded visualizations in Figures \tabref{fig:nextqa_supp_success}, \tabref{fig:star_supp_success}, \tabref{fig:perception_supp_success}, and abridged versions in \ref{fig:supp_nextqa_comparison} that demonstrate the benefits of our question-answering approach compared to regular captioning. For failures, we also present abridged versions for each dataset in Figures \tabref{fig:nextqa_supp_fail}, \ref{fig:star_supp_fail}, and \ref{fig:perception_supp_fail}. Finally, we present some additional success and failure cases in \ref{fig:vis_results}. For the visualizations, we display 3 iterations of the question answering process with 2 Extractor QA pairs each, compressing the output text by displaying the most important parts for brevity. For the abridged versions, we display the video on top, and the traversal order using the numbered orange circles. In the row beneath, we display the frames in the order they are selected, and display corresponding Extractor question-answer output in yellow and captions in gray.

\section{Licenses and Privacy}
\label{supp:datasets:Licenses}
The license, PII, and consent details of each dataset are in the respective papers. In addition, we wish to emphasize that the datasets we use do not contain any harmful or offensive content, as many other papers in the field also use them. Thus, we do not anticipate a specific negative impact, but, as with any Machine Learning method, we recommend to exercise caution.

\end{document}